\documentclass[lettersize,journal]{IEEEtran}
\usepackage{amsmath,amsfonts}
\usepackage{algorithmic}
\usepackage{algorithm}
\usepackage{array}
\usepackage[caption=false,font=normalsize,labelfont=sf,textfont=sf]{subfig}
\usepackage{textcomp}
\usepackage{stfloats}
\usepackage{url}
\usepackage{verbatim}
\usepackage{graphicx}
\usepackage{cite}
\usepackage{multirow}
\usepackage{booktabs}
\usepackage{makecell}
\usepackage{bbm}

\begin{document}

\newcommand{\sysname}{MINES}

\title{Message Intercommunication for Inductive Relation Reasoning}

\author{Ke Liang$^{\ast}$,
        Lingyuan Meng$^{\ast}$,
        Sihang Zhou,
        Siwei Wang,
        Wenxuan Tu,
        Yue Liu,
        Meng Liu, \\
        Xinwang Liu$^{\dag}$, ~\IEEEmembership{Senior~Member,~IEEE}

\IEEEcompsocitemizethanks{\IEEEcompsocthanksitem $^{\ast}$ Equal contribution.
\IEEEcompsocthanksitem $^{\dag}$ Corresponding Author.
\IEEEcompsocthanksitem Ke Liang, Lingyuan Meng, Xinwang Liu, Yue Liu, Meng Liu, Siwei Wang, and Wenxuan Tu are with the School of Computer, National University of Defense Technology, Changsha, 410073, China. E-mail: {xinwangliu@nudt.edu.cn}.
\IEEEcompsocthanksitem Sihang Zhou is with the College of Intelligence Science and Technology, National University of Defense Technology, Changsha, 410073, China.
}
\thanks{This work has been submitted to the IEEE for possible publication. Copyright may be transferred without notice, after which this version may no longer be accessible.}}

\markboth{Journal of \LaTeX\ Class Files,~Vol.~14, No.~8, August~2015}%
{Shell \MakeLowercase{\textit{et al.}}: Bare Advanced Demo of IEEEtran.cls for IEEE Computer Society Journals}


\maketitle


%












\begin{abstract}
Inductive relation reasoning for knowledge graphs, aiming to infer missing links between brand-new entities, has drawn increasing attention. The models developed based on Graph Inductive Learning, called GraIL-based models, have shown promising potential for this task. However, the uni-directional message-passing mechanism hinders such models from exploiting hidden mutual relations between entities in directed graphs. Besides, the enclosing subgraph extraction in most GraIL-based models restricts the model from extracting enough discriminative information for reasoning. Consequently, the expressive ability of these models is limited. To address the problems, we propose a novel GraIL-based inductive relation reasoning model, termed \sysname{}, by introducing a \underline{M}essage \underline{I}ntercommunication mechanism on the \underline{N}eighbor-\underline{E}nhanced \underline{S}ubgraph. Concretely, the message intercommunication mechanism is designed to capture the omitted hidden mutual information. It introduces bi-directed information interactions between connected entities by inserting an undirected/bi-directed GCN layer between uni-directed RGCN layers. Moreover, inspired by the success of involving more neighbors in other graph-based tasks, we extend the neighborhood area beyond the enclosing subgraph to enhance the information collection for inductive relation reasoning. Extensive experiments on twelve inductive benchmark datasets demonstrate that our \sysname{} outperforms existing state-of-the-art models, and show the effectiveness of our intercommunication mechanism and reasoning on the neighbor-enhanced subgraph.
\end{abstract}


\section{Introduction}
\begin{figure}[t]
\centering
\setlength{\abovecaptionskip}{-0.02cm}
\includegraphics[width=0.47\textwidth]{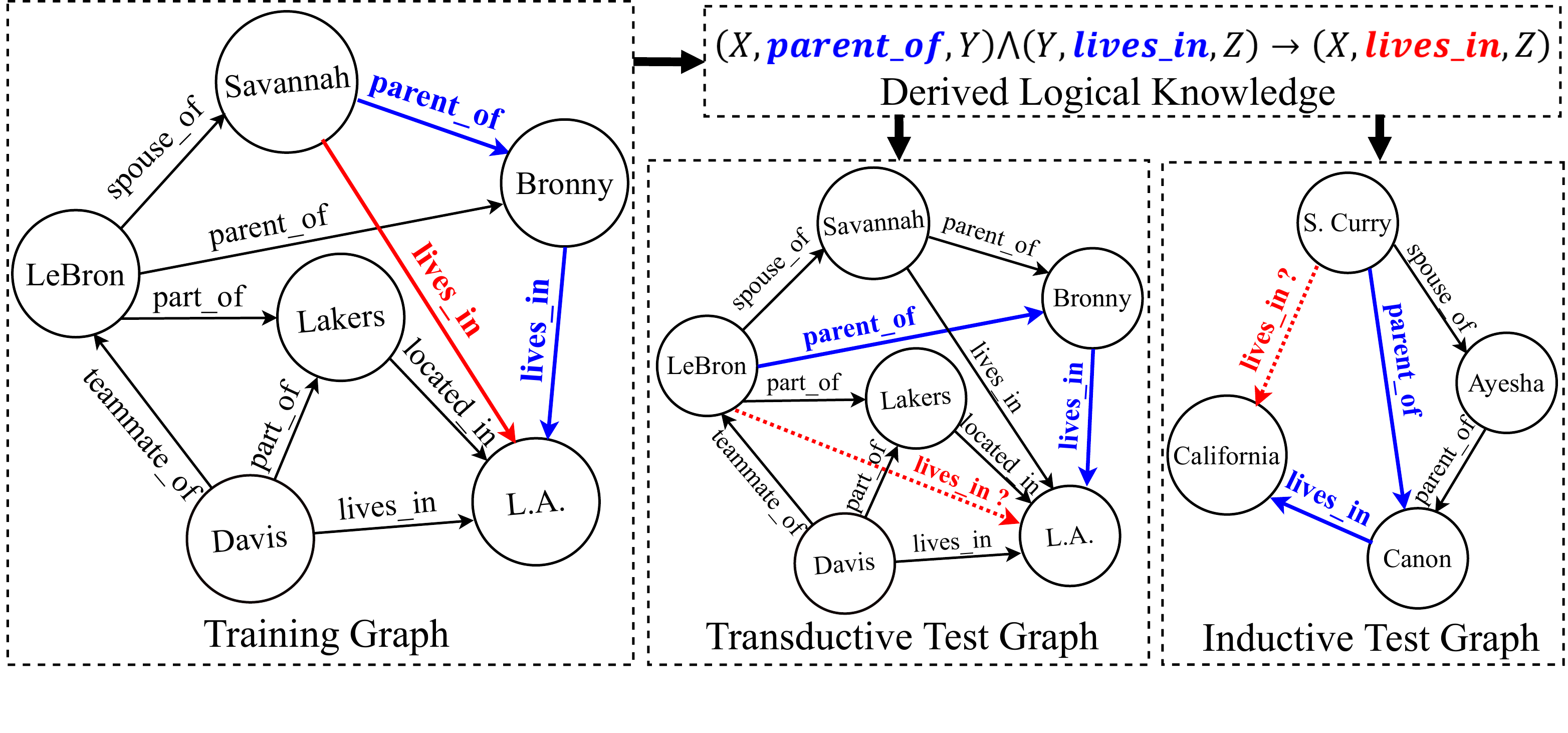}
\caption{Illustration of transductive and inductive relation reasoning. In the transductive scenario, entities in test graphs are all seen in the model during training. While as for the inductive scenario, unseen entities may exist in test graphs.}
\label{Illu_IR}  
\end{figure}

\IEEEPARstart{K}{nowledge} graphs (KGs) organize human knowledge in the form of the relational fact triplet. Each triplet consists of a head entity, a tail entity, and a relational edge between them. Recently, many applications have been developed based on KGs, such as dialogue generation \cite{liang2021mka, zhang2022subgraph}, recommendation systems \cite{wang2018ripplenet,liang2022reasoning,10.1145/3477495.3531985,10.1145/3459637.3481979}, code analysis \cite{liang2023abslearn}, and etc. However, most of the KGs, such as WikiData \cite{vrandevcic2012wikidata} and FreeBase \cite{bollacker2008freebase}, suffer from incompleteness issues. As an essential way to address the problem, relation reasoning, \textit{i.e.,} relation prediction, can be generally divided into two categories \cite{rossi2021knowledge, GraIL}, including transductive relation reasoning and inductive relation reasoning (See Fig. \ref{Illu_IR}). In fact, the inductive scenario is more common in the real world, \textit{e.g.,} new users (\textit{i.e.,} entities) are added in e-commerce KGs over continuous time \cite{trivedi2017know}. Therefore, more attention has recently been drawn to inductive models, which can infer the missing links between brand-new entities, and our research also falls into this category.

Rule-based and GNN-based methods are two typical inductive relation reasoning methods \cite{meng2023sarf}. Rule-based methods, such as NeuralLP \cite{NeuralLP}, RuleN \cite{RuleN}, and DRUM \cite{DRUM}, induce the entity-independent rules based on observed co-occurrence patterns. Such methods are naturally suitable for inductive scenarios with inherent inductive attributes. However, they suffer from limited expressive ability, and scalability \cite{GraIL}. Inspired by the great achievements of GNN-based methods for other graph-based tasks, several GNN-based inductive relation reasoning models have recently been proposed. Among them, Graph Inductive Learning \cite{GraIL}, \textit{i.e.,} GraIL, is the most influential. It first leverages RGCN \cite{RGCN} to infer missing triplets based on the enclosing subgraph, which gains great inductive ability. Based on the GraIL, many incremental GraIL-based models (\textit{e.g.,} TACT \cite{TACT}, CoMPILE \cite{ComPILE}, Meta-iKG \cite{Meta-iKG}, RPC-IR \cite{RPC-IR}, etc) are proposed, and all achieve promising performances.

\begin{figure}[t]
\centering
\includegraphics[width=0.47\textwidth]{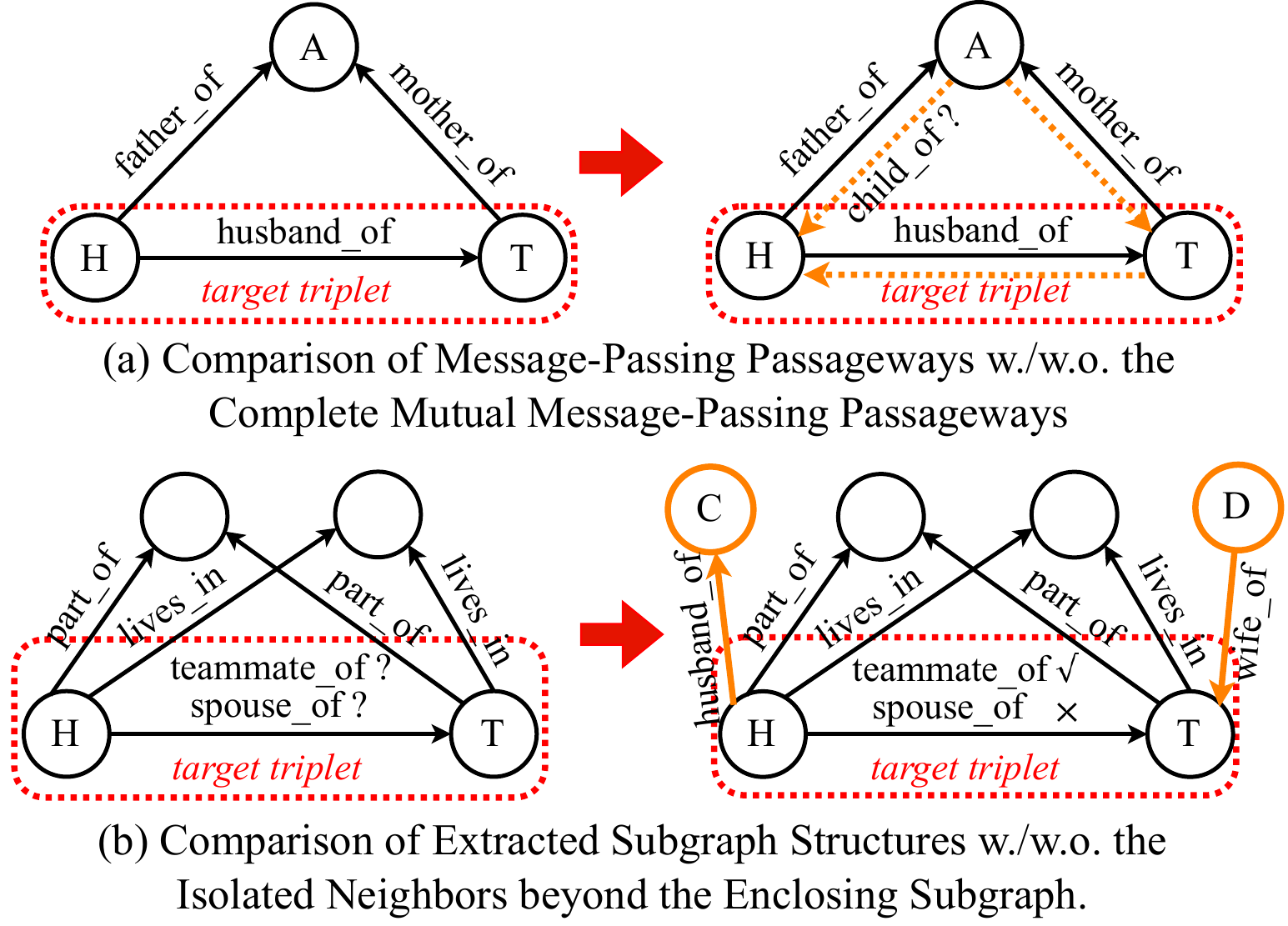}
\caption{Limitations of existing GraIL-based models. The differences between ideal scenarios (right figures) and scenarios in previous models (left figures) are colored in orange.}
\label{Prob}  
\end{figure}

Although proven effective, drawbacks still exist in GraIL-based models from two aspects. \textbf{(1)} \textbf{Insufficient message communication}: as we know, mutual relationships can always be found between two related entities in the real world. However, due to the incompleteness issue, such mutual relationships will usually not both exist in the given KG, \textit{e.g.,} the edge (\emph{A}, \emph{child\_of}, \emph{H}), which represents the mutual relationship corresponding to (\emph{H}, \emph{father\_of}, \emph{A}), does not exist in Fig. \ref{Prob} (a). Meanwhile, existing message-passing mechanisms can only aggregate messages along the given edges. Therefore, without such mutual relational edges, the expressive ability of the model is limited. However, we argue that the absence of edges in graphs does not mean the absence of message-passing passageways. For example, the orange passageways should exist in Fig. \ref{Prob} (a). Thus, we want a more powerful message communication mechanism by leveraging the omitted mutual information. \textbf{(2)} \textbf{Insufficient neighborhood information collection}: the existing GraIL-based models only perform reasoning on the enclosing subgraph composed of the paths between the target head and tail entities. Such subgraph extraction fashions abandon many neighbors around the target entities, which have been proven important in other tasks \cite{chen2022neighbor,liu2022local,niu2021relational}. In this work, we argue that isolated neighbors around the target entities beyond the enclosing subgraph will benefit the discriminative ability of the models. For example, the subgraph with the isolated neighbors, \textit{i.e.,} \emph{C} and \emph{D}, will be more informative for the model to distinguish the edges representing the \emph{teammate\_of} and \emph{spouse\_of} relations, which refers to the same enclosing subgraph structures in Fig. \ref{Prob} (b). Thus, performing reasoning based on a new subgraph with more isolated neighbors is worth a try.

Following the ideas above, we propose a novel GraIL-based inductive relation reasoning model, named \sysname{}, by introducing a \underline{M}essage \underline{I}ntercommunication mechanism on the \underline{N}eighbor-Enhanced \underline{S}ub-graph. Concretely, we first extract the neighbor-enhanced subgraph by including isolated neighbors around the target entities beyond the enclosing subgraph. Then, a sequential message intercommunication mechanism is designed to introduce bi-directed information interactions between connected entities. It is achieved by inserting an undirected/bi-directed graph convolutional network (GCN) layer between each of two uni-directed relational graphs convolutional network (RGCN) layers to compensate for the omitted hidden mutual information. Since we only know such missing mutual relational edges are very likely to exist but cannot tell the exact relation (\textit{i.e.,} edge label), the homogeneous subgraph without edge labels is used for intercommunication. In this way, our \sysname{} can achieve better expressive ability for inductive relation reasoning with acceptable computational cost proven by the experiments.
The main contributions are summarized as follows:

\begin{itemize}
    \item We propose a novel inductive relation reasoning model, \sysname{}, which effectively improves the expressive ability of GraIL-based models by introducing a message intercommunication mechanism on the neighbor-enhanced subgraph.

    \item  We innovatively design the message intercommunication mechanism. It introduces the bi-directed message interactions between connected entities to compensate for the omitted mutual relational information. Besides, we first perform inductive relation reasoning on the neighbor-enhanced subgraph to enhance the discriminative ability of models.
    
    \item Extensive experiments on twelve inductive datasets demonstrate that our \sysname{} outperforms existing state-of-the-art models, and show the effectiveness of the intercommunication mechanism and the strategy of reasoning on the neighbor-enhanced subgraph. Besides, we comprehensively analyze the impressive properties of our model from various aspects, including computation cost, intercommunication framework, transfer ability, transductive ability, etc.
\end{itemize}

\begin{figure*}[t]
\centering
\setlength{\abovecaptionskip}{-2pt}
\includegraphics[width=\textwidth]{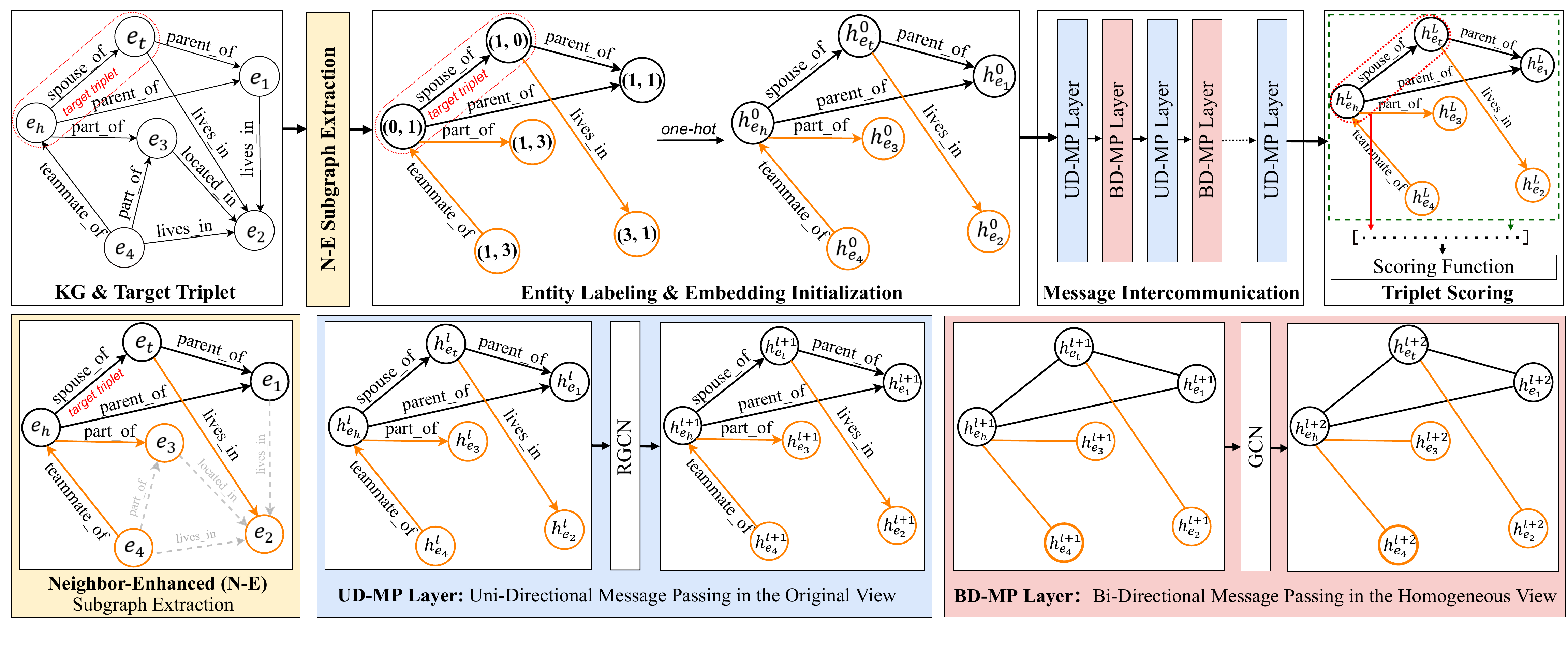}
\caption{The framework of the proposed \sysname{}. The framework includes four main steps: neighbor-enhanced subgraph extraction, entity labeling $\&$ embedding initialization, message intercommunication, and triplet scoring. Our main contribution lies in the first and the third step. Precisely, in the first step, we extract the neighbor-enhanced subgraph by adding neighboring entities and edges (colored in orange) to the enclosing subgraph (colored in black). In the previous work, only the black entities and relations are included in the subgraph for reasoning; In the third step, we feed subgraphs into the novel message intercommunication module to learn representations. 
In this step, GCN is integrated with RGCN to achieve better information interactions. Note that the \emph{1-hop} subgraph is taken as an example for the illustration.}
\label{OVERRALL_FIGURE}  
\end{figure*}

The paper is well organized as follows. The related work in Section II provides effective support to understand the motivation and innovation of this article. Section III comprehensively introduces the methodology of our model, \textit{i.e.,} \sysname{}. Then, we carefully describe the experiments, which aim to evaluate our model, in Section IV. Finally, we present a conclusion, as well as a discussion on future works, in Section V.

\section{RELATED WORK}\label{rw}
We divide this section into the following four parts, \textit{i.e.,} Transductive Relation Reasoning Methods, Inductive Relation Reasoning Methods, Message Passing in GNN-based Models, and Subgraph Extraction in GraIL-based Models. The first two parts summarize typical knowledge graph relation reasoning models, which constitute the compared baselines in this paper. The last two parts correspond to our motivations by discussing the limitations of previous GraIL-based models.

\subsection{Transductive Relation Reasoning Methods}
Transductive relation reasoning methods are usually embedding-based methods these years, including TransE \cite{TransE}, DisMult \cite{yang2015embedding} and their variants \cite{ji2015knowledge,trouillon2017knowledge,dettmers2018convolutional,sun2018rotate,cui2022reinforced,10.1145/3219819.3219986, SymclKGE}. However, these methods inherently assume a fixed entity set \cite{GraIL, yang2016revisiting}, which generally refers to transductive scenarios, instead of inductive scenarios. In general, transductive settings are the most used scenario in our real life, while it is not the best. Under this setting, models are usually handling tasks with unseen atoms, \textit{e.g.,} relations, entities, by retraining the models again to ensure visibility of the models \cite{GraIL}. It is really resource-consuming. Thus, researchers tend to study in inductive settings, which is also the scope we studied in this paper.

\subsection{Inductive Relation Reasoning Methods}
Rule-based and GNN-based methods are two typical inductive relation reasoning methods. Rule-based methods induce logical rules in KGs according to observed co-occurrences of frequent patterns. RuleN \cite{RuleN} and AMIE \cite{AMIE} set empirical thresholds based on the number of statistical results to mine the rules. Moreover, NeuralLP \cite{NeuralLP} and DRUM \cite{DRUM} derive rules in an end-to-end differentiable way. However, they suffer from limited expressive ability and scalability. Inspired by the great achievements of GNN for other graph-based tasks, several GNN-based inductive relation reasoning models have recently been proposed. The GNN models based on Graph Inductive Learning, \textit{i.e.,} GraIL-based models, \cite{GraIL} are the most influential among them. The prototype GraIL \cite{GraIL}, as the landmark GNN-based model, first leverages RGCN to perform the reasoning based on the local enclosing subgraph. Based on it, many incremental works are developed, including TACT \cite{TACT}, CoMPILE \cite{ComPILE}, Meta-iKG \cite{Meta-iKG}, RPC-IR \cite{RPC-IR}, and etc. These GraIL-based models all achieve promising inductive performances. Among them, TACT \cite{TACT} and CoMPILE \cite{ComPILE} both raise the importance of relation embeddings in the task. Concretely, TACT uses topology-aware correlations between relations to generate representations for triplet scoring. Besides, CoMPILE enhances the message interactions between relations and entities with a novel mechanism. After that, some popular strategies are also integrated, such as contrastive learning models (\textit{e.g.,} RPC-IR \cite{RPC-IR}, etc.) and meta-learning models (\textit{e.g.,} Meta-iKG \cite{Meta-iKG}, etc.). 
The above inductive relation reasoning models constitute our baselines.

\subsection{Message Passing in GNN-based Models}
The message-passing schemes aim to achieve information communication between entities, which are the basis of the GNN models \cite{liu2023self, liu2022hard, yue2022survey, liu2022embedding}. Based on the basic message-passing scheme proposed in vanilla GNN \cite{GNN_Book,scarselli2008graph}, various strategies are integrated to achieve better expressive ability, such as the GCN \cite{bruna2014spectral,welling2016semi} and GAT \cite{velivckovic2018graph}, etc. Then, with these ideas extended from homogeneous to other graph types, more message-passing schemes come out, including HAN \cite{wang2019heterogeneous}, GATNE \cite{cen2019representation}, HGNN \cite{feng2019hypergraph}, RGCN \cite{RGCN}, etc. However, the above message-passing mechanisms can only aggregate the message along the given edges. However, the mutual relational edges are usually missing in the given KGs, which restricts the expressive ability of the model. However, we argue that the absence of edges in graphs does not mean the absence of message-passing passageways. For example, the passageway from entity \emph{A} to \emph{T} corresponding to (\emph{T}, \emph{mother\_of}, \emph{A}) is supposed to exist in Fig. \ref{Prob} (a). Some works try to conquer this problem by directly adding “inverse relation” \cite{ChainACL, CompGCN} to provide such bi-directional message communication, which will sometimes lead to incorrect inversing edge construction. It will further hinder the discriminative ability of models. For example, the edge \emph{(A, FatherOf, B)} and its inversing edges \emph{(B, FatherOf, A)} can never both exist. To address the above concerns, a novel message intercommunication mechanism in our work is proposed.

\subsection{Subgraph Extraction in GraIL-based Models}
Recently, GraIL-based models have shown promising potential for the task. The prototype GraIL \cite{GraIL} first performs inductive relation reasoning on the undirected enclosing subgraph, which only considers the entities within the paths between the target entities. Like GraIL, its variants, \textit{i.e.,} other GraIL-based models, all perform reasoning on the enclosing subgraph, but some make specific implementation modifications on subgraph extraction. For example, CoMPILE \cite{ComPILE} extracts the directed enclosing subgraph instead of the undirected enclosing subgraph to improve the inference performance on symmetrical triplets. In our work, inspired by the success of exploiting more neighborhood information for GNN-based models for other tasks \cite{hamilton2017inductive,chen2022neighbor,liu2022local,niu2021relational, SNRI}, we notice that a certain number of isolated neighbors outside the enclosing subgraph will also benefit inductive relation, reasoning models. However, directly applying their subgraph strategies \cite{33333,44444} to our task will include more useless nodes in subgraphs, which may hinder the reasoning efficiency and accuracy. Compared to it, our \sysname{} is the first model to perform inductive relation reasoning on neighbor-enhanced subgraphs. Moreover, the filtering procedure for subgraph extraction described in Sec.3.2 effectively filters the useless nodes and improves the reasoning performance. 




\section{Method}
\subsection{Preliminary}
The knowledge graph is the directed relational graph, denoted as \emph{KG} $=(\mathcal{E}, \mathcal{R}, \mathcal{G})$, where the entity (\textit{i.e.,} node) set and the relation (i.e., edge label) set are represented as $\mathcal{E}$ and $\mathcal{R}$, respectively, and $\mathcal{G}=\{(e_u,r_{u,v},e_v)\ | \ e_u,e_v \in \mathcal{E}, r_{u,v} \in \mathcal{R}\}$ is the set of fact triplets (\textit{i.e.,} edges) in the given \emph{KG}. The main goal of the inductive relation reasoning is to predict the likelihood of the target relation $r_{t}$ between the target head $e_h$ and target tail $e_t$ by scoring the target triplet $(e_h,r_{t},e_t)$ in the given \emph{KG}. The notations are shown in Tab. \ref{NOTATION_TABLE}. Our \sysname{} is implemented based on the prototype GraIL \cite{GraIL}, and the main ideas in our paper have good scalability, which can be easily applied to other GraIL-based inductive models. For a fair comparison, we follow the settings in previous GraIL-based models to perform reasoning solely based on the structural semantics derived from the subgraph. The proposed \sysname{} have four steps (See Fig. \ref{OVERRALL_FIGURE}): {(1)} neighbor-enhanced subgraph extraction, {(2)} entity labeling and embedding initialization, {(3)} message intercommunication, and {(4)} triplet scoring.

\subsection{Neighbor-Enhanced Subgraph Extraction}
The neighbor-enhanced (N-E) subgraph in our \sysname{} is extracted by including more isolated neighbors beyond the paths between the target entities based on the enclosing subgraph extracted. Similar to \cite{GraIL}, we first generate the \emph{k-hop} neighbors around the target head and tail for both incoming and outgoing edges, denoted as $N_{k}(e_h)$ and $N_{k}(e_t)$. Then, we take the intersection of the neighbor sets and get the enclosing subgraph by filtering out the entities which are isolated beyond the paths between target entities. However, different from the enclosing subgraph in previous GraIL-based models, the neighbor-enhanced subgraph further contains the \emph{k-hop} isolated entities around the target head and target tail entities, together with the corresponding edges, which constitutes the \emph{k-hop} path between each isolate entity and the target entity (See \textbf{orange lines} in Fig. \ref{OVERRALL_FIGURE}). In this way, our \sysname{} enlarges the original enclosing subgraph to the more informative neighbor-enhanced subgraph for reasoning, which enhances the discriminative ability.
\begin{table}[t]
\centering
\fontsize{9.5}{11}\selectfont 
\caption{Notation summary}
\resizebox{\linewidth}{!}{
\begin{tabular}{cc}
\hline
{Notation} & {Explanation} \\\hline
  \emph{KG}      &    Knowledge graph    \\
$\mathcal{E}$ & Entity set in \emph{KG}\\
$\mathcal{R}$ & Relation set in \emph{KG}\\
$\mathcal{G}$ & Triplet set in \emph{KG}\\
$e_{\emph{i}}$ & Entity $i$\\
$r_{\emph{t}}$ & Relation $t$\\
\textbf{h}$_{\emph{e}}^l$ & Embedding vector of entity \emph{e} at $l^{th}$ layer\\
\emph{d}($e_i, e_j$) & distance between $e_i$ and $e_j$\\
${\bf{g}}_{ud}$($\cdot$)   &  Uni-directional message passing encoder\\ 
${\bf{g}}_{bd}$($\cdot$)    &  Bi-directional message passing encoder\\ 
$\mathcal{N}_{i}^{r}$   &  Entity set of neighbors with the relation $r$\\ 
${\bf{W}}_{r}^{l}$, ${\bf{W}}_{0}^{l}$ & Weight parameters \\
$\sigma(\cdot)$  & Activation function\\
$\gamma$ & Margin hyper-parameter\\
$|{\mathcal{E}_{NE}}|$  & The quantity of the entity elements\\
\textbf{h}$_{{\operatorname{KG}_{NE}}_{(e_h,r_t,e_t)}}$ & Subgraph representation\\
$L$ & Layers of network\\\
$\mathcal{L}$ & Loss function\\
\hline
\end{tabular}}
\label{NOTATION_TABLE} 
\end{table}

\subsection{Entity Labeling and Embedding  Initialization}
The entity features are required for the message communication mechanism in GNN models. Since no prior entity attributes are used in our framework, we leverage the structural semantics to initialize entity features, including two steps, \textit{i.e.,} entity labeling and embedding initialization. Concretely, each entity $e_i$ in the subgraph for the target triplet ($e_{h}, r_{t}, e_{t}$) is labeled with the distance tuple ($d(e_{i},e_{h}), d(e_{i},e_{t})$), where $d(e_{i},e_{h})$ represents the shortest distance of the undirected path between $e_i$ and $e_h$ without counting any path through $e_h$ (likewise for $d(e_{i},e_{t})$). Fig. \ref{OVERRALL_FIGURE} presents an example. Afterward, we initialize the embedding by the one-hot operation, denoted as ${\bf{h}}_{e_{i}}^{0}= [\operatorname{one-hot}(\emph{d}(e_{i},e_{h})) \oplus \operatorname{one-hot}(\emph{d}(e_{i},e_{t}))]$. Note that two target entities $e_h$ and $e_t$ are uniquely labeled as (0, 1) and (1, 0).

\subsection{Message Intercommunication Mechanism}
The message communication mechanisms in previous GraIL-based models for inductive relation reasoning mainly rely on the R-GCN \cite{RGCN}. With such a uni-directional message-passing scheme, the model will leave out the hidden mutual relational information underlying each edge in the given KG. Thus, the expressive ability of models is restricted without such information.

To achieve sufficient message communication, we propose a message intercommunication mechanism composed of the Uni-Directional Message Passing (\textit{i.e.,} \emph{UD-MP}) and Bi-Directional Message Passing (\textit{i.e.,} \emph{BD-MP}) layers, which are two different network layers with different message passing schemes. The message passing scheme in  \emph{UD-MP} layers is based on the original subgraph (\textit{i.e.,} original view), and the R-GCN is selected to update the embeddings as same as GraIL-based models. In comparison, the message passing scheme in the \emph{BD-MP} layer is selected as the simple GCN based on the homogeneous view of the subgraph. Besides, we use a sequential framework by inserting a \emph{UD-MP} layer (\textit{i.e.,} undirected GCN layer) between each of two directed \emph{BD-MP} layers (\textit{i.e.,} directed RGCN layers) in the proposed message intercommunication mechanism. The details will be illustrated as follows.

\subsubsection{Uni-Directional Message Passing in the Original View} 
We adopt the RGCN \cite{RGCN}, denoted as ${\bf{g}}_{ud}(\cdot)$, for embedding updating in the original view of the subgraph, which is the initially extracted subgraph without any modifications. The \emph{UD-MP} layers persevere the similar message communication schemes in previous GraIL-based models. They focus on the uni-directional message passing along the given edges in the KGs.
\begin{align}
    {\bf{h}}_{e_{i}}^{l+1} &={\bf{g}}_{ud}({\bf{h}}_{e_{i}}^{l})=\operatorname{\sigma}\left(\sum_{r\in\mathcal{R}}\sum_{e_{j}\in\mathcal{N}_{e_{i}}^{r}}\frac{1}{c_{e_{i},r}} {\bf{h}}_{e_{j}}^{l} {\bf{W}}_{r}^{l}+{\bf{h}}_{e_{i}}^{l} {\bf{W}}_{0}^{l}\right),
\end{align}where the feature vector of entity $e_{i}$ at the $l^{th}$ layer is present as ${\bf{h}}_{e_{i}}^{l}$. Besides, the set of neighbor indices for specific entity $e_{i}$ with the relation $r\in\mathcal{R}$ is marked as $\mathcal{N}_{i}^{r}$. $c_{e_{i},r}=|\mathcal{N}_{e_{i}}^{r}|$ is a normalization constant. ${\bf{W}}_{r}^{l}$, ${\bf{W}}_{0}^{l}$ are two weight parameters. Moreover, $\sigma(\cdot)$ is an activation function.

\subsubsection{Bi-Directional Message Passing in the Homogeneous View}
The basic GCN \cite{welling2016semi}, denoted as ${\bf{g}}_{bd}(\cdot)$, is adopted for embedding updating in the homogeneous view of the subgraph, which is generated by replacing the uni-directed labeled edges with the bi-directed/undirected unlabeled edges in the extracted subgraph (See Fig. \ref{OVERRALL_FIGURE}). 
Since we only know such missing mutual edges are very likely to exist but cannot tell the exact relation (\textit{i.e.,} edge label) without any language models, \textit{e.g.,} Bert \cite{devlin-etal-2019-bert}, we just build up the missing message-passing passageways with unlabeled edges in this work to bridge the message intercommunication between entities.
\begin{align}
    {\bf{h}}_{e_{i}}^{l+1} &={\bf{g}}_{bd}({\bf{h}}_{e_{i}}^{l})
    =\operatorname{\sigma}\left(\sum_{e_{j\in\mathcal{N}_{e_{i}}}}\frac{1}{c_{e_{i},e_{j}}} {\bf{h}}_{e_{j}}^{l}{\bf{W}}^{l}\right),
\end{align}where we leverage ${\bf{h}}_{e_{i}}^{l}$ to denote as the entity feature vector of $e_{i}$ at the $l^{th}$ layer. Besides, the set of neighbor indices of node $e_{i}$ is present as $\mathcal{N}_{i}$. Morever, a normalization constant for edge $(e_i,e_j)$ is calculated as  $c_{e_{i},e_{j}}=\sqrt{|\mathcal{N}_{e_{i}}|\cdot|\mathcal{N}_{e_{j}}|}$. Meanwhile, $\bf{W}^{l}$ and $\sigma(\cdot)$ denote the weight parameter and activation function separately.
\begin{figure}[t]
\centering
\includegraphics[width=0.47\textwidth]{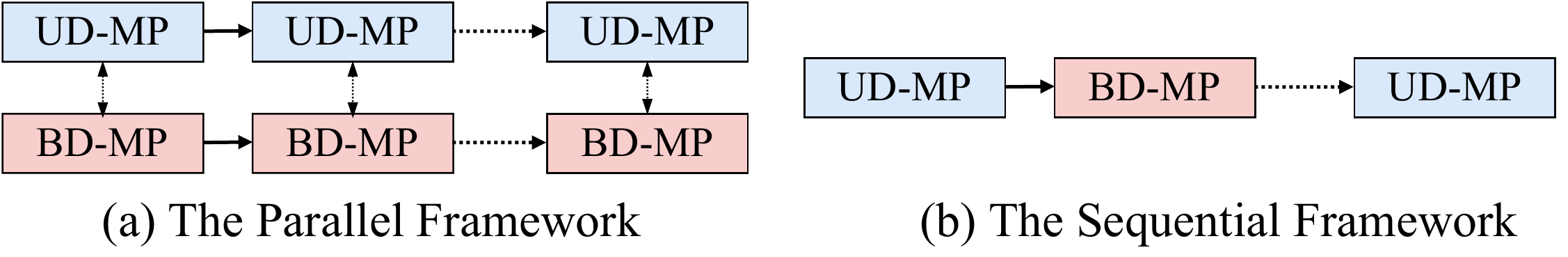}
\caption{Comparison of the parallel and sequential intercommunication frameworks.}
\label{paseq}  
\end{figure}

\subsubsection{Framework of the Intercommunication Mechanism} 
The sequential framework is used to integrate the \emph{UD-MP} and \emph{BD-MP} layers in our message intercommunication mechanism. Concretely, we insert a \emph{UD-MP} layer (\textit{i.e.,} undirected GCN layer) between each of two directed \emph{BD-MP} layers (\textit{i.e.,} directed RGCN layers) to compensate for the hidden mutual information omitted by each RGCN layer. Besides, we reassign the entity embedding generated in one view to the corresponding entity in the other view for the cross-view communication of the entity embeddings. Moreover, our model fixes the first and last network layer as the \emph{UD-MP} layer. The main reason for selecting a sequential framework instead of a parallel one is to reduce the complexity of the model. As shown in Fig. \ref{paseq}, parallel frameworks will generally take more GNN layers than sequential ones. Meanwhile, the substitution of the simpler GCN layer for the RGCN layer will lead to a reduction in the number of parameters. 

\subsection{Triplet Scoring}
The scoring function $f(e_h,r_t,e_t)$, which aims to measure whether the inferring is of high possibility or not, is calculated as follows in our \sysname{}.
\begin{align}
f(e_h,r_t,e_t)={\bf{W}}[{\bf{h}}_{KG_{{NE}_{(e_h,r_{t},e_t)}}}\oplus{\bf{h}}_{e_{h}}\oplus{\bf{h}}_{e_{t}}\oplus{\bf{h}}_{{r_t}}],
\end{align}where $\bf{W}$ denotes as the weight parameter, $\bf{h}_{e_{h}}$ and $\bf{h}_{e_{t}}$ represent the hidden embeddings of entities for head and tail respectively, the learned embedding of the target relation is marked as $\bf{h}_{r_{t}}$. $\bf{h}_{{KG_{NE}}_{(e_h,r_t,e_t)}}$, as the subgraph representation, is calculated as follows:
\begin{align}
    {\bf{h}}_{\operatorname{KG}_{{NE}_{(e_h,r_t,e_t)}}} =\frac{1}{|\mathcal{E}_{NS}|}\sum_{{e_i}\in {\mathcal{E}_{NE}}} \mathbf{h}_{e_i}^{L},
\end{align}where $|{\mathcal{E}_{NE}}|$ is the quantity of the entity elements in the set, and $L$ represents the quantity of the network layer used in the model, \textit{i.e.,} the sum of the quantity of \emph{UD-MP} layers and \emph{BD-MP} layers.

Based on the scoring function, we train the model to score positive triplets higher than the negative triplets via the noise-contrastive hinge loss \cite{TransE}:
\begin{align}
\mathcal{L}=\sum_{(e_h,r_t,e_t)\in \mathcal{G}}{\max (0, f(e^{n}_h,r^{n}_t,e^{n}_t)-f(e_h,r_t,e_t)+\gamma)},
\end{align}where $\gamma$ is the margin hyper-parameter, and we use ($e_h,r_t,e_t$) and ($e^{n}_h,r^{n}_t,e^{n}_t$) to represent the positive and negative triplets separately. In particular, we generate a negative triplet by replacing the head (or tail).

\section{Experiment}
In this section, the experimental settings are first carefully introduced. Then, we present the performance comparison based on the experiments, which demonstrate the superiority of our \sysname{} for inductive relation reasoning. Afterward, the ablation study shows the effectiveness of the message intercommunication mechanism and the strategy of reasoning on the neighbor-enhanced subgraph. Later, the transfer experiments illustrate that our ideas can be easily applied to other GraIL-based models. Moreover, we further comprehensively analyze the impressive properties of our model from various aspects, including intuitive case study, key parameter \emph{k}, and transductive ability.

\begin{table*}[t]
\fontsize{8}{13}\selectfont 
\caption{AUC-PR results on the inductive benchmark datasets. Best results are boldfaced, and the second best ones are underlined.}
\resizebox{\linewidth}{!}{
\begin{tabular}{cclcccccccccccccc}
\hline
\multicolumn{3}{c}{\multirow{2}{*}{Methods}}                                      & \multicolumn{4}{c}{WN18RR}    &  & \multicolumn{4}{c}{FB15K-237} &  & \multicolumn{4}{c}{NELL-995}  \\ \cline{4-7} \cline{9-12} \cline{14-17} 
\multicolumn{3}{c}{}                                                              & v1    & v2    & v3    & v4    &  & v1    & v2    & v3    & v4    &  & v1   & v2    & v3    & v4    \\ \hline
\multicolumn{1}{c}{\multirow{3}{*}{Rule-Based}}  &
\multicolumn{2}{c}{Neural-LP} & 86.02 & 83.78 & 62.90 & 82.06 &  & 69.64 & 76.55 & 73.95 & 75.74 &  & 64.66 & 83.61 & 87.58 & 85.69 \\
\multicolumn{1}{c}{}                             & \multicolumn{2}{c}{DRUM}      & 86.02 & 84.05 & 63.20 & 82.06 &  & 69.71 & 76.44 & 74.03 & 76.20 &  & 59.86 & 83.99 & 87.71 & 85.94 \\
\multicolumn{1}{c}{}                             & \multicolumn{2}{c}{RuleN}     & 90.26 & 89.01 & 76.46 & 71.75 &  & 75.24 & 88.70 & 91.24 & 91.79 &  & 84.99 & 88.40 & 87.20 & 80.52 \\ \hline
\multicolumn{1}{c}{\multirow{4}{*}{GraIL-based}} & \multicolumn{2}{c}{GralL}     & 94.32 & 94.18 & 85.80 & 92.72 &  & 84.69 & 90.57 & 91.68 & 94.46 &  & 86.05 & 92.62 & 93.34 & 87.50 \\
\multicolumn{1}{c}{}

&\multicolumn{2}{c}{TACT}     & {94.64} & {97.45}  & {86.33}  & {97.97}  &  & {83.82}  & {92.98}  & {91.28}  & {94.42}  &  & \underline{88.72}  & {94.80}  & {94.79}  & {85.76}  \\

\multicolumn{1}{c}{}  
& \multicolumn{2}{c}{CoMPILE}   & 98.23 & \textbf{99.56} & 93.60 & \textbf{99.80} &  & 85.50 & 91.68 & 93.12 & \underline{94.90} &  & 80.16 & \underline{95.88} & 96.08 & 85.48 \\
\multicolumn{1}{c}{}                             & \multicolumn{2}{c}{Meta-iKG}  & --     & --     & --     & --     &  & 80.31 & 82.95 & 82.52 & 84.23 &  & 72.12 & 84.11 & 82.47 & 79.25 \\
\multicolumn{1}{c}{}                             & \multicolumn{2}{c}{RPC-IR}    & \underline{99.41} & 93.76 & \underline{98.75} & 87.24 &  & \underline{92.75} & \underline{93.93} & \underline{95.26} & 84.23 &  & {88.12} & 94.12 & \underline{96.10}  & \underline{87.81} \\
\hline
\multicolumn{1}{c}{Ours}                         & \multicolumn{2}{c}{MINES}     & \textbf{99.69} & \underline{99.48} & \textbf{99.27} & \underline{99.58} &  & \textbf{99.01} & \textbf{99.41} & \textbf{99.56} & \textbf{99.48} &  & \textbf{99.55} & \textbf{99.59} & \textbf{99.70} & \textbf{97.52} \\ \hline
\end{tabular}
}
\label{LP_AUCPR}
\vspace{-0.2 cm}
\end{table*}
\begin{table}[t]
\fontsize{24}{44}\selectfont 
\caption{Inductive relation reasoning benchmark datasets. We represent the number of relations, entities, and fact triplets as \#R, \#E, and \#Tr, respectively. The train-graph and test-graph donate the knowledge graph (KG) for training and the new KG with unseen entities for evaluation, separately.}
\resizebox{\linewidth}{!}{
\begin{tabular}{lcccccccccc}
\hline
& \multicolumn{1}{l}{} & \multicolumn{3}{c}{WN18RR} & \multicolumn{3}{c}{FB15K-237} & \multicolumn{3}{c}{NELL-995} \\ \cline{3-11} 
&      & \#R    & \#E     & \#Tr    & \#R     & \#E      & \#Tr     & \#R   & {\#E}   & \#Tr   \\ \hline
\multirow{2}{*}{v1} & train-graph                & 9      & 2746    & 6678    & 183     & 2000     & 5226     & 14    & 10915       & 5540   \\
& test-graph                & 9      & 922     & 1991    & 146     & 1500     & 2404     & 14    & 225         & 1034   \\ \hline
\multirow{2}{*}{v2} & train-graph               & 10     & 6954    & 18968   & 203     & 3000     & 12085    & 88    & 2564        & 10109  \\
& test-graph                & 10     & 2923    & 4863    & 176     & 2000     & 5092     & 79    & 4937        & 5521   \\ \hline
\multirow{2}{*}{v3} & train-graph                   & 11     & 12078   & 32150   & 218     & 4000     & 22394    & 142   & 4647        & 20117  \\
& test-graph                  & 11     & 5084    & 7470    & 187     & 3000     & 9137     & 122   & 4921        & 9668   \\ \hline
\multirow{2}{*}{v4} & train-graph                & 9      & 3861    & 9842    & 222     & 5000     & 33916    & 77    & 2092        & 9289   \\
& test-graph                 & 9      & 7208    & 15157   & 204     & 3500     & 14554    & 61    & 3294        & 8520   \\ \hline
\end{tabular}
}
\vspace{-0.3 cm}
\label{dataset}
\end{table}

\subsection{Experiment Setting}
\subsubsection{Dataset} Most of the existing KG datasets, such as WN18RR \cite{dettmers2018convolutional}, NELL-995 \cite{NELL-995}, Yago \cite{yago,yago2,yago3,yago4}, FB15K-237 \cite{FB15K-237}, and DBPedia \cite{dbpedia}, are originally created for transductive relation reasoning. To well evaluate the inductive ability of the model, GraIL \cite{GraIL} first generates twelve benchmark datasets based on the FB15K-237, NELL-995, and WN18RR. In the datasets, no entity appears both in the training graphs and test graphs, and all of the relations in the test graph will appear in the corresponding training graph (See details in Tab. \ref{dataset}). These datasets are then widely used to evaluate the inductive relation reasoning models. In this work, we also use these twelve inductive datasets to evaluate our model with the existing state-of-the-art baselines. 
\begin{table}[t]
\fontsize{7}{11}\selectfont 
\caption{Hit@10 results on the WN18RR. Best results are boldfaced, and second best ones are underlined.}
\resizebox{\linewidth}{!}{
\begin{tabular}{cclcccc}
\hline
\multicolumn{3}{c}{\multirow{2}{*}{Methods}}                   & \multicolumn{4}{c}{WN18RR}                                        \\ \cline{4-7} 
\multicolumn{3}{c}{}                                           & v1             & v2             & v3             & v4             \\ \hline
\multirow{3}{*}{Rule-Based}    & \multicolumn{2}{c}{Neural-LP} & 74.37          & 68.93          & 46.18          & 67.13          \\
& \multicolumn{2}{c}{DRUM}      & 74.37          & 68.93          & 46.18          & 67.13          \\
& \multicolumn{2}{c}{RuleN}     & 80.85          & 78.23          & 53.39          & 71.59          \\ \hline
\multirow{5}{*}{GraIL-based}   & \multicolumn{2}{c}{GraIL}     & 82.45          & 78.68          & 58.43          & 73.41          \\
& \multicolumn{2}{c}{TACT}      & 83.24          & \underline{81.63}    & \underline{62.73}    & 76.27          \\
& \multicolumn{2}{c}{CoMPILE}   & 83.60          & 79.82          & 60.69          & 75.49          \\
& \multicolumn{2}{c}{Meta-iKG}  & --             & --             & --             & --             \\
& \multicolumn{2}{c}{RPC-IR}    & \underline{85.11}    & \underline{81.63}    & 62.40  & \underline{76.35}          \\ \hline
Ours                           & \multicolumn{2}{c}{MINES}  & \textbf{87.23} & \textbf{83.87} & \textbf{69.42} & \textbf{79.04} \\ \hline
\end{tabular}
}
\label{LP_HIt1}
\vspace{0.4 cm}
\fontsize{7}{11}\selectfont
\caption{Hit@10 results on the FB15K-237 and NELL-995. Best results are boldfaced, and second best ones are underlined.}
\resizebox{\linewidth}{!}{
\begin{tabular}{cclcccc}
\hline
\multicolumn{3}{c}{\multirow{2}{*}{Methods}}                   & \multicolumn{2}{c}{FB15K-237}   &\multicolumn{2}{c}{NELL-995}    \\ \cline{4-7} 
\multicolumn{3}{c}{}                                           & v1             & v2                            & v3             & v4             \\ \hline
\multirow{3}{*}{Rule-Based}    & \multicolumn{2}{c}{Neural-LP} & 52.92          & 58.94             & 82.71          & 80.58          \\
& \multicolumn{2}{c}{DRUM}      & 52.92          & 58.73              & 82.71          & 80.58          \\
& \multicolumn{2}{c}{RuleN}     & 49.76          & 77.82                     & 77.26          & 61.35          \\ \hline
\multirow{5}{*}{GraIL-based}   & \multicolumn{2}{c}{GraIL}     & 64.15          & 81.80                     & 91.41          & 73.19          \\
& \multicolumn{2}{c}{TACT}      & 65.61          & \underline{83.05}      & {91.35}    & 74.69          \\
& \multicolumn{2}{c}{CoMPILE}   & \underline{67.64}    & 82.98                      & 92.77          & \underline{75.19}    \\
& \multicolumn{2}{c}{Meta-iKG}  & 66.52          & 72.37                   & 77.99          & 71.63          \\
& \multicolumn{2}{c}{RPC-IR}    & {67.56}    & {82.53}       & \underline{94.01}    & 71.82          \\ \hline
Ours                           & \multicolumn{2}{c}{MINES}  & \textbf{67.67} & \textbf{83.18} &    \textbf{95.92} & \textbf{81.61} \\ \hline
\end{tabular}
}
\label{LP_HIt2}
\end{table}

\subsubsection{Implementation}
We implement our \sysname{} based on the prototype GraIL model \cite{GraIL} with the PyTorch library \cite{paszke2019pytorch} and conduct all experiments based on a single 12 GB NVIDIA TITAN XP. We select the 3-layer model (\textit{i.e.,} \emph{UD-MP}+\emph{BD-MP}+\emph{UD-MP}) and \emph{3-hop} extracted subgraphs as same as the previous GraIL-based models to fairly compare with SOTA models. Besides, the dimension of the feature representation and dropout rate are set to 32 and 0.5 separately. As for the optimizer for training, we use Adam \cite{ADAM} with the learning rate set to 0.001. Moreover, the batch size and the margin parameter $\gamma$ are set to 16 and 10 separately. The model is evaluated based on ranking and classification metrics (\textit{i.e.,} Hits@10 and AUC-PR). To calculate the area under the precision-recall curve (AUC-PR), we score an equal number of negative triplets sampled by replacing the head (or tail) with a random entity along with the triplet. As for Hits@10, we rank each test triplet among 50 other randomly sampled negative triplets. We report the mean results of five times experiments as previous works do.

\subsubsection{Compared Baselines}
Our model is compared with {typical inductive relation reasoning models} without using any context attributes, including Neural-LP \cite{NeuralLP}, DRUM \cite{DRUM}, RuleN \cite{RuleN}, GraIL \cite{GraIL}) and emerging GraIL-based models, including TACT \cite{TACT}, ComPILE \cite{ComPILE}, RPC-IR \cite{RPC-IR}, Meta-iKG \cite{Meta-iKG}. Note that we reproduce the results for TACT using the same methods for negative sampling as other GraIL-based models, and other results are obtained from the original paper.
\begin{figure*}[t]
\centering
\includegraphics[width=\textwidth]{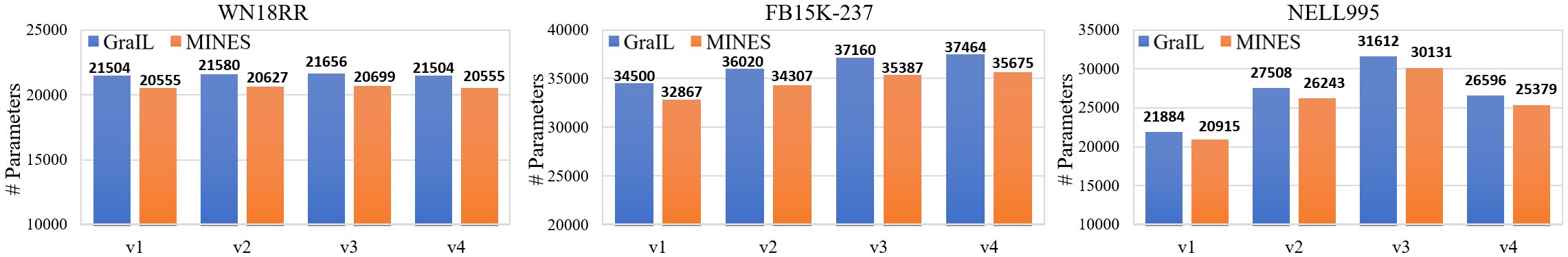}
\caption{Comparison of the parameter numbers of 3-layer \sysname{} and prototype GraIL for training on benchmark datasets.}
\label{Parameter}  
\end{figure*}
\begin{table*}[!htb]
\fontsize{8}{13}\selectfont 
\caption{AUC-PR results on the inductive benchmark datasets w./w.o. M.I. and N.E.. $\uparrow$ represents the value is increased compared to the baseline}
\resizebox{\linewidth}{!}{
\begin{tabular}{cllllcllllcllll}
\hline
\multicolumn{1}{c}{\multirow{2}{*}{Methods}}        & \multicolumn{4}{c}{\multirow{1}{*}{WN18RR}}     &\multirow{1}{*}{}                                                                & \multicolumn{4}{c}{\multirow{1}{*}{FB15K-237}}   &\multirow{1}{*}{}                                                                & \multicolumn{4}{c}{\multirow{1}{*}{NELL-995}}                                        \\ \cline{2-5}\cline{7-10}\cline{12-15}

\multicolumn{1}{c}{}                                & \multicolumn{1}{c}{\multirow{1}{*}{v1}} & \multicolumn{1}{c}{\multirow{1}{*}{v2}} & \multicolumn{1}{c}{\multirow{1}{*}{v3}} & \multicolumn{1}{c}{\multirow{1}{*}{v4}}&\multicolumn{1}{c}{\multirow{1}{*}{}} & \multicolumn{1}{c}{\multirow{1}{*}{v1}} & \multicolumn{1}{c}{\multirow{1}{*}{v2}} & \multicolumn{1}{c}{\multirow{1}{*}{v3}} & \multicolumn{1}{c}{\multirow{1}{*}{v4}}&\multicolumn{1}{c}{\multirow{1}{*}{}} & \multicolumn{1}{c}{\multirow{1}{*}{v1}} & \multicolumn{1}{c}{\multirow{1}{*}{v2}} & \multicolumn{1}{c}{\multirow{1}{*}{v3}} & \multicolumn{1}{c}{\multirow{1}{*}{v4}} \\
\hline

\multicolumn{1}{c}{\multirow{1}{*}{GraIL}}                & \multirow{1}{*}{94.32}&\multirow{1}{*}{94.18}&\multirow{1}{*}{85.80}& \multirow{1}{*}{92.72}& \multicolumn{1}{c}{\multirow{1}{*}{}}&
\multirow{1}{*}{84.69}&\multirow{1}{*}{90.57}&\multirow{1}{*}{91.68}& \multirow{1}{*}{94.46}& \multicolumn{1}{c}{\multirow{1}{*}{}}& \multirow{1}{*}{86.05}&\multirow{1}{*}{92.62}&\multirow{1}{*}{93.34}& \multicolumn{1}{c}{\multirow{1}{*}{87.50}}   \\

\multicolumn{1}{c}{\multirow{1}{*}{GraIL w. M.I.}}              & \multirow{1}{*}{99.50{$\uparrow$}}&\multirow{1}{*}{99.43{$\uparrow$}}&\multirow{1}{*}{99.17{$\uparrow$}}& \multirow{1}{*}{99.56{$\uparrow$}}& \multicolumn{1}{c}{\multirow{1}{*}{}}&
\multirow{1}{*}{98.46{$\uparrow$}}&\multirow{1}{*}{99.08{$\uparrow$}}&\multirow{1}{*}{99.52{$\uparrow$}}& \multirow{1}{*}{99.41{$\uparrow$}}& \multicolumn{1}{c}{\multirow{1}{*}{}}& \multirow{1}{*}{97.16{$\uparrow$}}&\multirow{1}{*}{99.50{$\uparrow$}}&\multirow{1}{*}{99.56{$\uparrow$}}& \multicolumn{1}{c}{\multirow{1}{*}{93.83}{$\uparrow$}}    \\       

\multicolumn{1}{c}{\multirow{1}{*}{GraIL w. N.E.}}              & \multirow{1}{*}{95.68{$\uparrow$}}&\multirow{1}{*}{95.66{$\uparrow$}}&\multirow{1}{*}{88.80{$\uparrow$}}& \multirow{1}{*}{95.22{$\uparrow$}}& \multicolumn{1}{c}{\multirow{1}{*}{}}&
\multirow{1}{*}{85.38{$\uparrow$}}&\multirow{1}{*}{90.94{$\uparrow$}}&\multirow{1}{*}{91.71{$\uparrow$}}& \multirow{1}{*}{94.69{$\uparrow$}}& \multicolumn{1}{c}{\multirow{1}{*}{}}& \multirow{1}{*}{87.80{$\uparrow$}}&\multirow{1}{*}{94.38{$\uparrow$}}&\multirow{1}{*}{94.21{$\uparrow$}}& \multicolumn{1}{c}{\multirow{1}{*}{89.81}{$\uparrow$}}    \\

\multicolumn{1}{c}{\multirow{1}{*}{\sysname{}}}                & \multirow{1}{*}{99.69{$\uparrow$}}&\multirow{1}{*}{99.48{$\uparrow$}}&\multirow{1}{*}{99.27{$\uparrow$}}& \multirow{1}{*}{99.58{$\uparrow$}}& \multicolumn{1}{c}{\multirow{1}{*}{}}&
\multirow{1}{*}{99.01{$\uparrow$}}&\multirow{1}{*}{99.41{$\uparrow$}}&\multirow{1}{*}{99.56{$\uparrow$}}& \multirow{1}{*}{99.48{$\uparrow$}}& \multicolumn{1}{c}{\multirow{1}{*}{}}& \multirow{1}{*}{99.55{$\uparrow$}}&\multirow{1}{*}{99.59{$\uparrow$}}&\multirow{1}{*}{99.70{$\uparrow$}}& \multicolumn{1}{c}{\multirow{1}{*}{97.52}{$\uparrow$}}    \\
\hline
\end{tabular}
}
\label{asauc}
\end{table*}

\subsection{Performance Comparison with Baselines}
Tab. \ref{LP_AUCPR}, Tab. \ref{LP_HIt1} and Tab. \ref{LP_HIt2} show that our \sysname{} significantly outperforms other compared baselines on inductive datasets for both the Hits@10 and AUC-PR evaluation metrics. On average, our method makes 4.01\% on AUC-PR and 2.78\% on Hit@10 boosts on each dataset compared to the previous best performances. Specifically, \sysname{} improves the best AUC-PR performance by an average of 6.06\% on NELL-995 and FB15K-237 datasets. It further highlights the better discriminative and expressive ability of our \sysname{} with the novel message intercommunication mechanism and novel strategy of reasoning on the neighbor-enhanced subgraph. Besides, we observe that the improvement of our Hit@10 performance on FB15K-237 is not apparent, which indicates that our model may be more effective for the sparser datasets with fewer relations. Besides, Fig. \ref{Parameter} shows that our \sysname{} is a lightweight model compared to the prototype GraIL model. For each dataset, our \sysname{} reduces about 1500 parameters for training. Such parameter reduction is caused by replacing an RGCN layer with a simpler GCN layer. Thus, the above results demonstrate the superiority of \sysname{} from both the evaluation metrics and model complexity aspects. 
\begin{table}[t]
\fontsize{5}{8}\selectfont 
\caption{Hit@10 results on WN18RR w./w.o. M.I. and N.E.. $\uparrow$ represent the value increased compared to the baseline.}
\resizebox{\linewidth}{!}{
\begin{tabular}{cllll}
 \hline
& \multicolumn{4}{c}{WN18RR}                       \\ \cline{2-5} 
\multirow{-2}{*}{Methods} & \multicolumn{1}{c}{v1}   & \multicolumn{1}{c}{v2} & \multicolumn{1}{c}{v3}   & \multicolumn{1}{c}{v4}\\ \hline
GraIL             &   82.45&    78.68                                        & 58.43 & 73.41       \\
GraIL w. M.I.            & {84.04{$\uparrow$}} & {80.27{$\uparrow$}}                                    & {60.99{$\uparrow$}} & {75.79{$\uparrow$}}    \\ 
GraIL w. N.E.     &   87.03{$\uparrow$}&    82.35{$\uparrow$}                                         & 68.21{$\uparrow$}   &  78.90{$\uparrow$}       \\
MINES   &    {87.23{$\uparrow$}}      &{83.87{$\uparrow$}}                     &    {69.42{$\uparrow$}}      &{79.04{$\uparrow$}}    \\ \hline
\end{tabular}
}
\label{ashit1}
\vspace{0.3 cm}
\fontsize{5}{8}\selectfont 
\caption{Hit@10 results on the FB15K-237 and NELL-995 w./w.o. M.I. and N.E.. $\uparrow$ represent the value increased compared to the baseline.}
\resizebox{\linewidth}{!}{
\begin{tabular}{cllll}
 \hline
& \multicolumn{2}{c}{FB15K-237}  & \multicolumn{2}{c}{NELL-995}                      \\ \cline{2-5} 
\multirow{-2}{*}{Methods} & \multicolumn{1}{c}{v1}   & \multicolumn{1}{c}{v2} & \multicolumn{1}{c}{v3}   & \multicolumn{1}{c}{v4}\\ \hline
GraIL             & 64.15         & 81.80         & 91.41         & 73.19      \\
GraIL w. M.I.            & 64.90 {$\uparrow$}        & 82.43 {$\uparrow$}        & 93.33 {$\uparrow$}       & 77.62 {$\uparrow$}   \\ 
GraIL w. N.E.     & 66.29   {$\uparrow$}      & 82.53    {$\uparrow$}     & 91.41    {}    & 75.81  {$\uparrow$}    \\
MINES      & 67.67  {$\uparrow$}       & 83.18 {$\uparrow$}        & 95.92   {$\uparrow$}      & 81.61  {$\uparrow$}   \\ \hline
\end{tabular}}
\label{ashit2}
\end{table}

\subsection{Ablation Study}
The ablation studies are performed on multiple benchmarks to investigate the effectiveness and robustness of the strategy of reasoning on the neighbor-enhanced (N.E.) subgraph and message intercommunication (M.I.) mechanism in our \sysname{}. Two compared models (\textit{i.e.,} GraIL w. M.I. and GraIL w. N.E.) are generated. In GraIL w. M.I., only the message intercommunication mechanism is integrated with the GraIL. In GraIL w. N.E., the original uni-directional message communication mechanism is kept, but the neighbor-enhanced subgraph substitutes the enclosing subgraph. Tab. \ref{asauc} shows that the average AUC-PR values on WN18RR, FB15K-237, and NELL-995 increased by 7.66\%, 8.77\%, and 7.64\% with the M.I. mechanism, which is higher compared to improvements of the neighbor-enhanced subgraph (\textit{i.e.,} 2.01\%, 0.33\%, 1.67\%) on these datasets. It suggests that the message intercommunication mechanism is more effective for classification performance. Besides, Tab. \ref{ashit1} and Tab. \ref{ashit2} show that the average Hit@10 values are increased by 1.98\% with the M.I. mechanism, while lower than the average improvement of 3.62\% brought by leveraging the N.E. procedure. It indicates that the ranking performance benefits more from the N.E. subgraph. Based on promising results, the effectiveness and robustness of each module have been proven.
\begin{figure*}[t]
\centering
\includegraphics[width=\textwidth]{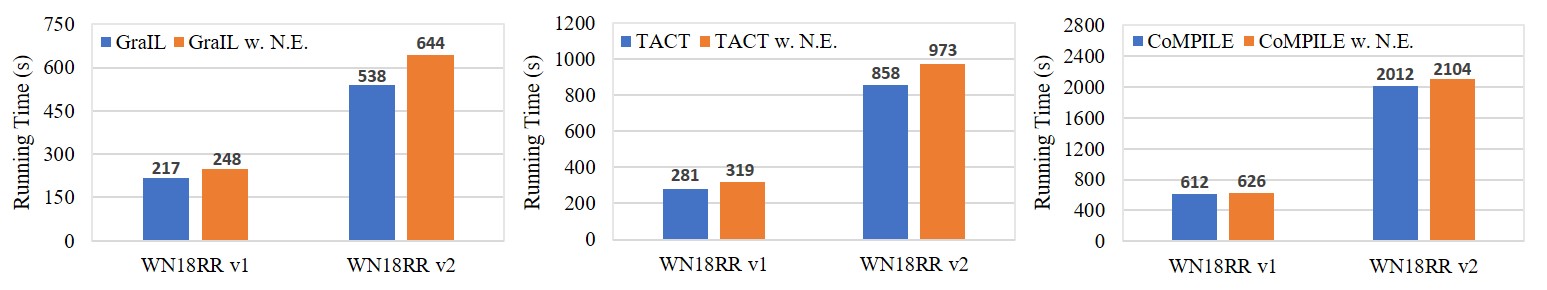}
\caption{Computational cost analysis on GraIL-based models w./w.o. neighbor-enhanced subgraph extraction module.}
\label{runningtime}  
\end{figure*}

\subsection{Subgraph Extraction Computation Analysis} It is easy to understand that collecting more information from neighbors can enhance the expressive ability of the GNN-based models, which is proven in the previous section. However, considering more entities will also increase the computational costs. Fig. \ref{runningtime} shows the computational cost comparison of the GraIL-based models w./w.o. neighbor-enhanced (N.E.) subgraph extraction. Concretely, GraIL, TACT, and CoMPILE are selected as the baselines with the hop \emph{k} set to 3, and the experiments are conducted on two inductive benchmarks, WN18RR v1, and WN18RR v2. We observe that the computation overhead raised by performing reasoning on neighbor-enhanced subgraphs is still acceptable. The computational cost is positively correlated with the number of entities. Since limited entities will be added in our neighbor-enhanced subgraph, \textit{\textit{i.e.,}} only k-hop neighbors are added around the head and tail entity, the running time increase is also limited theoretically. The experimental results also prove it. 
\begin{table}[!t]
\fontsize{28}{47}\selectfont
\caption{Performance comparison between different intercommunication frameworks. The best results are boldfaced.}
\resizebox{\linewidth}{!}{
\begin{tabular}{ccccccc}
\hline
& \multicolumn{2}{c}{WN18RR v1}                                 & \multicolumn{2}{c}{FB15K-237 v1}                              & \multicolumn{2}{c}{NELL-955 v1}                                                         \\ \cline{2-7} 
\multirow{-2}{*}{Methods} & AUC-PR               & { {Hit@10}} & AUC-PR & { {Hit@10}}& AUC-PR& {{Hit@10}} \\ \hline
RRR                    & 95.68          & 87.03                                  & 84.69          & 64.15                                  & 86.05          & 59.50                                   \\
Bi-RRR                    & 95.96          & 82.45                                  & 83.10           & 62.44                                  & 70.64          & 58.50                                   \\
GGG                       & 97.00             & 84.04                                  & 82.99          & 61.22                                  & 67.11          & 47.50                                \\
GRR                       & 96.02          & 84.04                                  & 85.61          & 64.15                                  & 86.04          & 60.50                                   \\
RRG                       & 99.50           & 84.04                                  & 98.78          & 54.63                                  & 99.51          & 54.50                                   \\ \hline
RGR (Ours)                      & \textbf{99.69} & \textbf{87.23}                         & \textbf{99.01} & \textbf{67.67}                         & \textbf{99.55} & \textbf{63.50}                          \\ \hline
\end{tabular}
}
\label{AIF}
\end{table}

\subsection{Intercommunication Framework Analysis} To demonstrate the suitability of the proposed sequential intercommunication framework, we investigate various combinations of the \emph{BD-MP} and \emph{UD-MP} layers in our \sysname{} on three benchmark datasets. Concretely, the compared frameworks include (1) RRR, (2) Bi-RRR, (3) GGG, (4) GRR, (5) RRG, and (6) RRR (\textit{i.e.,} the baseline without any inter-communication frameworks), where R represents one \emph{UD-MP} layer, \textit{i.e.,} one RGCN layer in the original view of the subgraph, and G represents one \emph{BD-MP} layer, \textit{i.e.,} one GCN layer in the homogeneous view of the subgraph. In particular, Bi-RRR is a 3-RGCN-layer model in the bi-directional subgraph by adding all of the inversing edges to the original subgraph. Tab. \ref{AIF} shows that the RGR used in \sysname{} outperforms Bi-RRR and GGG on average by 16.63\% on AUC-PR and 6.78\% on Hit@10. It indicates that the negative impacts of the redundant relation information in the Bi-RRR and the relationship information loss in GGG influence the performance more than the positive impacts of the intercommunication. Besides, compared to GRR and RRG, the average AUC-PR and Hit@10 are higher by 5.17\% and 5.82\%. It suggests that fixing the first and last RGCN layers benefits the discriminative ability of the models for the relation reasoning tasks in KGs. In conclusion, our RGR is the proper framework.
\begin{table}[t]
\fontsize{9}{13}\selectfont 
\caption{Transfer experiments on TACT and CoMPILE. The best results are boldfaced.}
\resizebox{\linewidth}{!}{
\begin{tabular}{ccccc}
\hline
& \multicolumn{2}{c}{WN18RR v1}                       & \multicolumn{2}{c}{NELL-955 v1}                          \\ \cline{2-5} 
\multirow{-2}{*}{Methods} & AUC-PR   & Hit@10 & AUC-PR    & Hit@10 \\ \hline
GraIL             &   94.32&    82.45                                         & 86.05 & 59.50       \\
GraIL 
w. M.I.N.E.            & \textbf{99.69} & \textbf{87.23}                                    & \textbf{99.55} & \textbf{63.50}    \\ \hline
TACT             &   94.64&    83.24                                         & 85.58 & 55.00       \\
TACT 
w. M.I.N.E.            & \textbf{95.55} & \textbf{85.11}                                    & \textbf{88.72} & \textbf{62.00}    \\ \hline
CoMPILE      &   98.23&    83.60                                          & 80.16   &  58.38       \\
CoMPILE
w. M.I.N.E.       &    \textbf{100.00}      &\textbf{87.50}                     &    \textbf{100.00}      &\textbf{60.00}    \\ \hline
\end{tabular}
}
\label{extf}
\end{table}

\subsection{Transfer Analysis on TACT and CoMPILE}
The results in previous sections have shown that our strategies can benefit the prototype GraIL model. In this section, we further extend our idea to TACT and CoMPILE, two typical GraIL-based models, to evaluate the scalability and generalizability of our approach. The new models (\textit{i.e.,} TACT w. M.I.N.E. and CoMPLIE w. M.I.N.E.) are implemented by replacing the subgraph extraction and message communication modules with the neighbor-enhanced subgraph extraction and message intercommunication mechanism in our \sysname{}. Tab. \ref{extf} shows the significant improvements in both AUC-PR and Hit@10 metrics on two benchmark datasets (\textit{i.e.,} WN18RR v1 and NELL-995 v1) for both of the new models. It demonstrates that the ideas in \sysname{} can be well-scaled to other GraIL-based models.
\begin{figure*}[!t] 
\centering
\includegraphics[width=\textwidth]{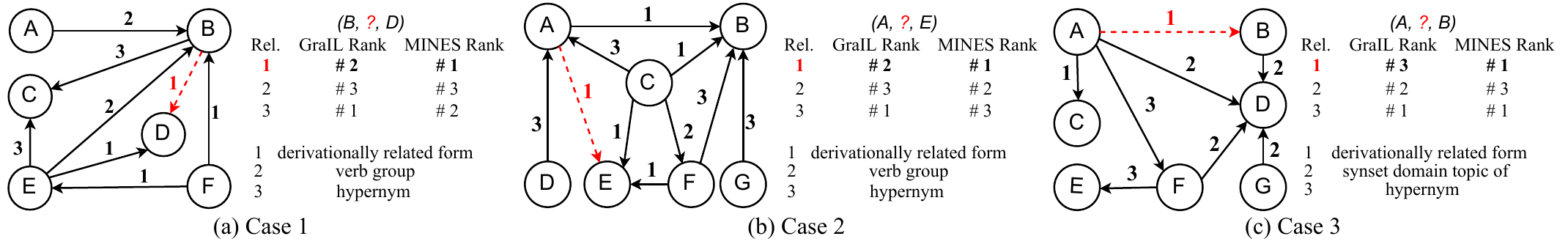}
\caption{Intuitive case study of \sysname{} and the prototype GraIL on the real-world cases derived from WN18RR v1}
\label{case}  
\end{figure*}

\subsection{Parameter Analysis on Hop $k$}
\begin{figure}[t]
\centering
\includegraphics[width=0.47\textwidth]{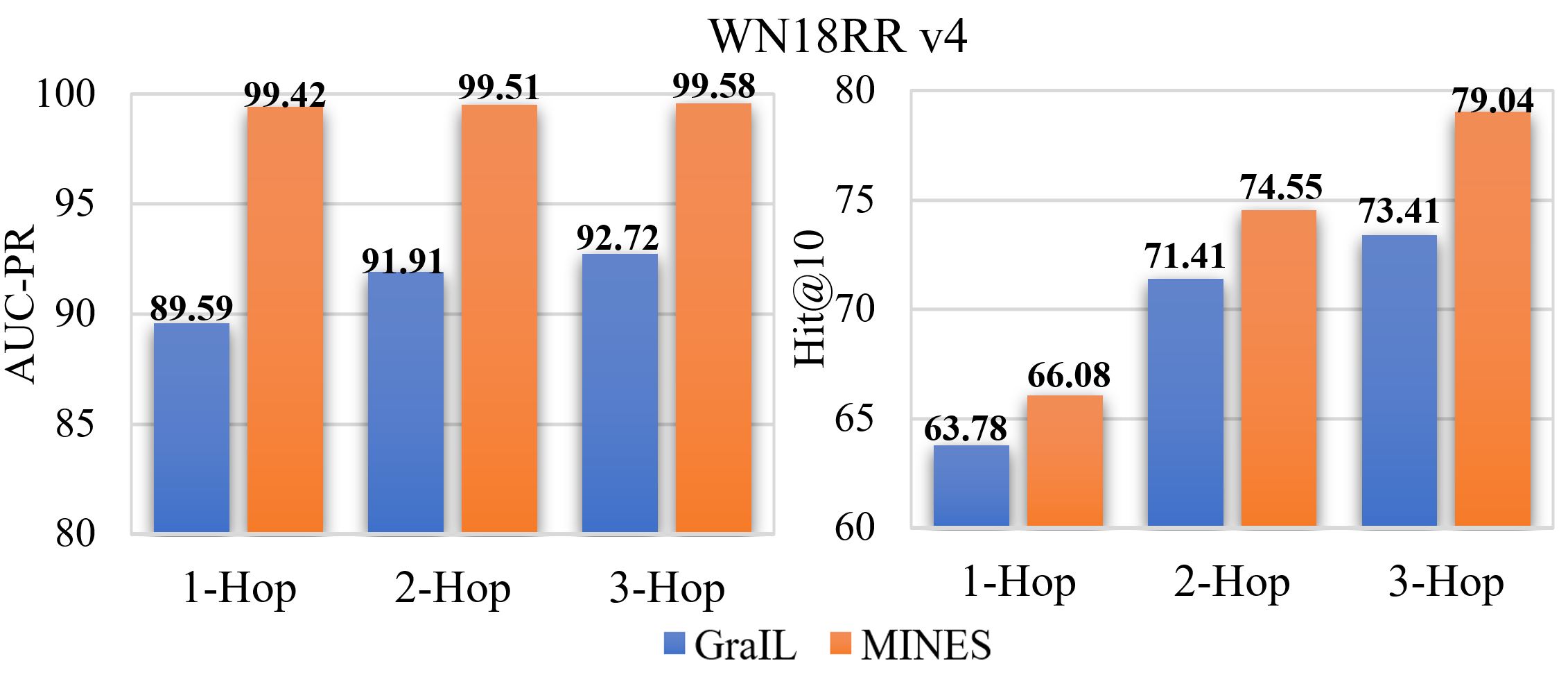}
\caption{Parameter analysis on hop \emph{k} of \sysname{} compared with the prototype GraIL.}
\label{stat_non}  
\end{figure}

The experiments on parameter hop $k\in\{1,2,3\}$ are performed on the WN18RR v4 to analyze how it influences the performance. Fig. \ref{stat_non} shows that the ranking metric Hit@10 is more sensitive to $k$ than the classification metric AUC-PR, which is similar to GraIL. Meanwhile, our \sysname{} can get an average of 8.09\%, and 3.69\% boosts on AUC-PR and Hit@10 with different hops compared to prototype GraIL. It further demonstrates the excellent scalability of our model with different hops.

\subsection{Intuitive Case Study}
We test the trained \sysname{} and GraIL on three toy cases derived from WN18RR v1, where $k=2$. Fig. \ref{case} shows that \sysname{} can get higher ranks for target triplets between the target entities. As for the triplet (\emph{A}, \emph{derivationally related form}, \emph{B}) in Fig. \ref{case} (c), the messages for both \emph{A} and \emph{B} will not get any messages from other entities except themselves with the prototype GraIL based on 1-hop enclosing subgraph. However, the intercommunication between the entities happens within our \sysname{}, which makes message interaction more sufficient (likewise for the cases in Fig. \ref{case} (a) and Fig. \ref{case} (b)). As a result, the higher ranks generated by our model intuitively demonstrate a better discriminative ability.
\begin{table}[t]
\fontsize{6}{8}\selectfont
\caption{Transductive ability analysis on \sysname{} on WN18RR compared to some typical transductive and inductive models. The best results are boldfaced.}
\resizebox{\linewidth}{!}{
\begin{tabular}{cccc}
\hline
\multicolumn{2}{c}{}                                 & \multicolumn{2}{c}{WN18RR}                                               \\ \cline{3-4} 
\multicolumn{2}{c}{\multirow{-2}{*}{Methods}}        & {{ Hit@1}} & {{Hit@10}} \\ \hline
& TransE       & 0.021                              & 0.533                               \\
& RotatE       & 0.428                              & 0.571                               \\
& QuatE        & 0.440                               & 0.551                               \\
& HAKE         & 0.453                              & 0.582                               \\
& DisMult      & 0.370                               & 0.521                               \\
& ComplEX      & 0.445                              & 0.574                               \\
& RGCN      & 0.382                              & 0.510                               \\
& SCAN      & 0.430                              & 0.540  \\
& KBGAT      & 0.426                              & 0.539                               \\
\multirow{-8}{*}{Transductive Model} & CompGCN      & 0.443                              & 0.546                               \\
& CURL      & 0.429                              & 0.523                               \\
& SymCL-KGE      & 0.429                              & 0.585                               \\
\hline
& Neural LP    & 0.371                              & 0.566                               \\
& DRUM         & 0.425                              & 0.586                               \\
& GraIL        &    0.671                                & 0.730                         \\
& ComPILE        &    0.673                                & 0.742                        \\
\multirow{-5}{*}{Inductive Model}     & MINES (Ours) & \textbf{0.678}                              & \textbf{0.768}                               \\ \hline
\end{tabular}}
\vspace{-0.2 cm}
\label{TranR}
\end{table}

\subsection{Transductive Ability Analysis}
As inductive relation reasoning requires a more generalized expressive ability of the model, the inductive models should perform well on transductive relation reasoning. To prove that, we evaluate the transductive ability of our model compared with several existing state-of-the-art models on transductive relation reasoning, including TransE \cite{TransE}, RotatE \cite{sun2018rotate}, QuatE \cite{zhang2019quaternion}, HAKE \cite{zhang2020learning}, DisMult \cite{yang2015embedding}, ComplEX \cite{zhang2020duality}, RGCN \cite{RGCN}, CURL \cite{CURL}, SCAN \cite{SCAN}, KBGAT \cite{KBGAT}, CompGCN \cite{CompGCN}, and SymCL-KGE \cite{SymclKGE}. Some of the inductive models are also evaluated, including NeuralLP \cite{NeuralLP}, DRUM \cite{DRUM}, GraIL \cite{GraIL} and ComPILE \cite{ComPILE}. Restricted by the GPU memory, the experiments are conducted on WN18-RR \cite{dettmers2018convolutional}, which contains more than 40000 entities and 90000 edges. Tab. \ref{TranR} shows that our \sysname{} outperforms other traditional transductive models by large margins, \textit{i.e.,} about 25\% and 19\% improvements compared to other state-of-the-art models. The experimental results are as expected, the transductive ability of our \sysname{} is also promising.

\section{Conclusion}
In this paper, we propose a novel GraIL-based inductive relation reasoning model, named \sysname{}, by introducing a Message Intercommunication mechanism on the Neighbor-Enhanced Subgraph. As a result, our model is of better discriminative and expressive ability due to sufficient information communication. Extensive experiments on twelve inductive benchmark datasets demonstrate that our \sysname{} outperforms existing state-of-the-art models and show the effectiveness of our message intercommunication mechanism and the strategy of reasoning on the neighbor-enhanced subgraph. Besides, a comprehensive analysis is presented to illustrate the impressive properties of our model from various aspects. 

In the future, we aim to investigate neighbor-enhanced subgraphs for inductive relation reasoning in a more fine-grained and efficient manner. Besides, we are also excited to explore transferring the message intercommunication mechanism to enhance the message communication procedures in other GNN-based models for better expressive ability. Moreover, as an effective inductive relation reasoning model, we also tend to apply our model to some specific AI applications, such as drug structure inference, recommendation systems, etc.

\bibliographystyle{IEEEtran}
\bibliography{sample-base}

\end{document}